\documentclass[11pt]{article}

\usepackage[preprint]{acl}

\usepackage{times}
\usepackage{latexsym}

\usepackage[T1]{fontenc}

\usepackage[utf8]{inputenc}

\usepackage{microtype}

\usepackage{inconsolata}

\usepackage{graphicx}

\usepackage{amsmath}
\usepackage{amssymb}
\usepackage{booktabs}
\usepackage{multirow}
\usepackage{algorithm}
\usepackage{algpseudocode}
\usepackage{enumitem}
\usepackage{hyperref}
\usepackage{url}
\usepackage{xcolor}

\title{CA-RAG: Constraint-Aligned Retrieval for Structurally Faithful Long-Form Generation}

\author{
  \textbf{K. Vimal Kumar} \quad
  \textbf{Meghana Kshirsagar} \quad
  \textbf{Conor Ryan} \\
  Lero -- The Research Ireland Centre for Software \\
  Department of Computer Science \& Information Systems \\
  University of Limerick, Limerick, Ireland
}

\begin{document}
\maketitle
\begin{abstract}
Retrieval-augmented generation (RAG) typically optimises evidence selection
for semantic relevance, whereas compliance-sensitive long-form generation
also requires evidence supporting explicit structural obligations.
Semantically relevant retrieval may therefore omit information needed for
mandatory downstream clauses. We propose Constraint-Aligned RAG
(\textsc{CA-RAG}), a retrieval-conditioning framework that augments
section-specific queries with ontology-derived structural constraints and
policy-grounded knowledge-graph relations before evidence selection.
Retrieval-only evaluation with manually annotated evidence shows
consistent improvements across retrieval budgets. At $k=6$, \textsc{CA-RAG}
increases Precision@$k$ from $0.5268$ to $0.5692$, pooled Recall@$k$ from
$0.5402$ to $0.5950$, and F1@$k$ from $0.5389$ to $0.5672$ over standard
RAG. Generation experiments further show improvements in compliance and
reliability in several model--domain settings, while generation-only
constraints achieve higher clause coverage but lower overall compliance and
reliability. These results indicate that structurally aligned retrieval can improve
evidence sufficiency in compliance-sensitive long-form generation. 
\end{abstract}

\section{Introduction}

Large language models (LLMs) have demonstrated strong capabilities in conditional text generation and few-shot adaptation on a wide range of natural language processing tasks \citep{brown2020language}. However, many real-world long-form generation tasks, including contracts, policy reports, clinical documentation, regulatory disclosures, and enterprise workflow artefacts, require outputs that satisfy explicit structural and compliance constraints in addition to factual relevance. In these settings, generation quality depends not only on retrieving semantically related evidence, but also on preserving clause-level completeness and document-level consistency across extended contexts.

Retrieval-augmented generation (RAG) improves grounded generation by incorporating external evidence during decoding \cite{lewis2021retrieval}. However, existing RAG pipelines optimise retrieval primarily for semantic similarity to the input query, largely independent of the structural requirements that govern the final document \cite{zhao2024retrieval,gao2024rag}. Semantically relevant retrieval may still omit structurally required evidence, leading to omission failures, inconsistent sections, and incomplete policy synthesis in long-form generation settings. We hypothesise that a substantial subset of these failures arises from structurally incomplete retrieval rather than decoding alone.

Recent work has explored integrating structured knowledge graphs (KG) and relational reasoning into language models to improve grounding and interpretability \cite{yasunaga2022qa}. However, these approaches primarily focus on reasoning and answer selection rather than retrieval for long-form generation. Similarly, existing query expansion, reranking, and prompt-based control methods improve semantic recall or generation control without explicitly conditioning retrieval on downstream structural requirements.

We propose Constraint-Aligned RAG (CA-RAG), a retrieval-conditioning framework that incorporates ontology-derived schema constraints and policy-grounded relational dependencies prior to evidence selection. Unlike query expansion or reranking, CA-RAG does not simply improve semantic recall or reorder retrieved passages. Instead, structural constraints condition which evidence becomes retrievable for downstream generation. Ontology constraints encode document-level structure, while knowledge graph relations capture dependencies between clauses, entities, and policy conditions. By aligning retrieval with downstream requirements, CA-RAG aims to improve structural completeness and reduces omission failures in long-form generation.

We evaluate \textsc{CA-RAG} across education, finance, and healthcare
using three open-weight LLMs and multiple retrieval baselines. We additionally evaluate retrieval independently using manually judged evidence to isolate retrieval-stage effects. Results show improved conventional retrieval performance and domain-dependent changes in structural evidence sufficiency, with downstream gains in compliance and reliability across several model--domain settings. These findings highlight the value of structurally sufficient evidence for compliance-sensitive long-form generation.

\noindent\textbf{Contributions.}
Our contributions are:
\begin{enumerate}[itemsep=1pt, topsep=1pt]
\item We identify a \emph{retrieval--generation mismatch}: semantically relevant evidence may be structurally insufficient for mandatory content.
\item We propose \textsc{CA-RAG}, which conditions retrieval on ontology-derived requirements and policy-grounded KG dependencies.
\item We evaluate retrieval independently using manually annotated evidence, standard information retrieval (IR) metrics, and structural evidence measures.
\item Across three domains and three LLMs, we show improved evidence recovery and domain-dependent gains in structural sufficiency, with corresponding downstream improvements in several settings.

\end{enumerate}

\section{Related Work}

\noindent\textbf{Retrieval-Augmented Generation.}
Retrieval-augmented generation (RAG) improves grounded text generation by conditioning language models on external evidence retrieved at inference time \cite{lewis2021retrieval}. Subsequent work has extended RAG across adaptive retrieval strategies, retrieval architectures, and knowledge-intensive tasks, including question answering, summarisation, and long-context generation \cite{zhao2024retrieval,gao2024rag}. However, most RAG systems optimise retrieval primarily for semantic similarity to the input query, without explicitly considering whether the retrieved evidence is structurally sufficient for downstream long-form generation. Consequently, semantically relevant retrieval may still produce incomplete evidence for section-wise synthesis and clause-level consistency.

\noindent\textbf{Controlled, Knowledge-Grounded, and Adaptive Retrieval Methods.}
A complementary line of work explores structure-aware and controllable generation through hierarchical planning, dynamic content control, and constrained decoding \cite{hu2022planet,you2023eipe,stolfo2025}. Other approaches integrate structured knowledge graphs and relational reasoning to improve factual grounding and interpretability \cite{yasunaga2022qa,li2024retrollm}. Adaptive retrieval methods such as \textsc{Self-RAG} \cite{selfrag} and \textsc{Flare-RAG} \cite{flarerag} further refine retrieval dynamically during generation to improve factual consistency and response quality.

Despite these advances, existing approaches generally target semantic relevance, evidence refinement, relational grounding, or generation-time control rather than explicitly conditioning retrieval on the structural requirements of the target document. Query expansion methods typically introduce semantically related terms, KG-based retrieval incorporates structured relational knowledge, and re-ranking methods reorder candidate passages using downstream relevance signals. In contrast, CA-RAG conditions retrieval using ontology-derived and policy-grounded structural constraints before evidence selection, explicitly guiding retrieval toward structurally sufficient evidence for downstream generation.

\noindent\textbf{Positioning of This Work.}
CA-RAG therefore shifts the focus from improving semantic relevance alone to enforcing structural sufficiency during retrieval itself. This directly targets omission failures arising from incomplete retrieval by constraining evidence selection before generation begins. 

\section{Methodology}

\begin{figure*}[t]
  \centering
  \includegraphics[width=0.9\linewidth]{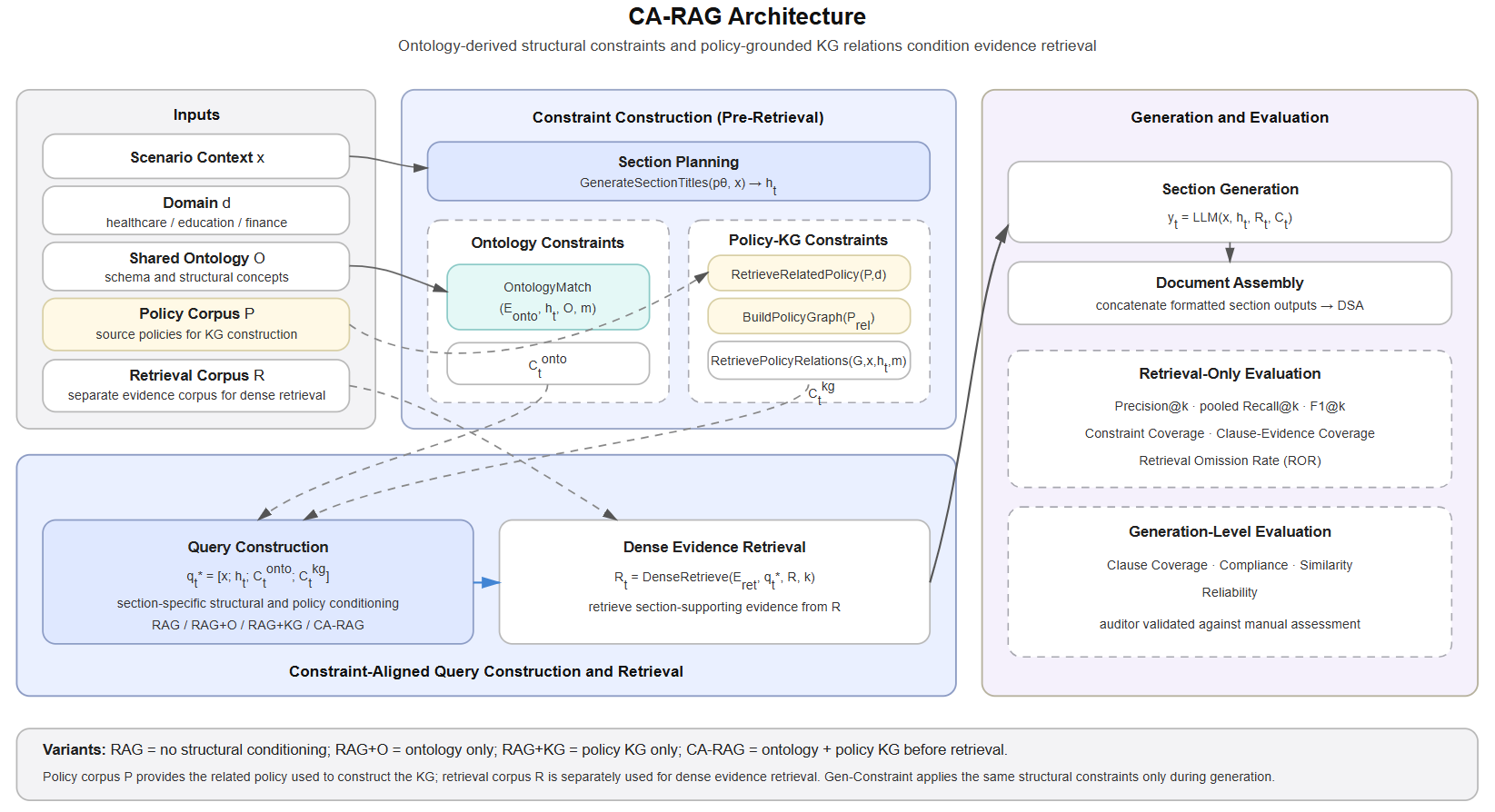}
  \caption{Overview of \textsc{CA-RAG}: retrieval conditioned on structural and policy constraints for long-form generation. }
  \label{fig:system}
\end{figure*}

\subsection{Problem Formulation}

We study long-form structured generation, where document validity depends
not only on semantic relevance, but also on satisfying explicit structural and policy constraints.

Given an input scenario $x$ and an external corpus $\mathcal{R}$, the goal is to generate a structured document
$
y=(y_1,\dots,y_T)
$
composed of multiple sections. Semantically relevant retrieval may still omit structurally required evidence, leading to omission failures, incomplete policy synthesis, and inconsistent sections. Our focus is on fluency, structural completeness, consistency, and grounding. 

Let $p_\theta(y \mid c)$ denote a language model conditioned on context $c$.
Standard RAG defines
\begin{equation}
c=(x,R(x)),
\end{equation}
where retrieval is optimised primarily for semantic similarity. We instead condition retrieval using structured constraints:
\begin{equation}
\label{eq:carag}
c=\big(x,R(x \mid \mathcal{C}(x)),\mathcal{C}(x)\big),
\end{equation}
where $\mathcal{C}(x)$ denotes ontology-derived and policy-grounded structural constraints.

CA-RAG is a retrieval-conditioning framework rather than a new retriever architecture. It can be viewed as structurally grounded query conditioning: rather than expanding queries with semantically related terms alone, CA-RAG derives additional retrieval terms from ontology-defined document requirements and policy-grounded KG relations.

\subsection{Baseline Retrieval-Augmented Generation}

Our baseline follows a standard dense RAG pipeline. Given scenario context $x$, the retriever selects evidence
\begin{equation}
R_x \subset \mathcal{R},
\end{equation}
and generation proceeds as
\begin{equation}
\label{eq:llm}
y=\mathrm{LLM}(x,R_x).
\end{equation}

\subsection{Constraint-Aligned RAG}

CA-RAG conditions evidence retrieval using ontology-derived schema constraints and policy-grounded relational dependencies extracted before retrieval.

\paragraph{Ontology Constraints.}
The ontology layer captures schema-level structure, including section types, mandatory fields, and attribute dependencies. For each section heading $h_t$, relevant ontology concepts are retrieved through semantic matching:
\begin{equation}
\mathcal{C}^{\text{onto}}_t=\mathrm{Match}(h_t,\{o_l\}).
\end{equation}

\paragraph{Policy-Grounded KG Constraints.}
We additionally construct policy-grounded knowledge graphs whose nodes represent clauses, entities, and policy attributes, while edges encode relations such as \textsc{requires}, \textsc{permits}, and \textsc{prohibits}. Relevant subgraphs are retrieved as
\begin{equation}
\mathcal{C}^{\text{kg}}_t=\mathrm{RetrieveKG}(x,h_t).
\end{equation}

The final constraint representation is
\begin{equation}
\mathcal{C}_t=\mathcal{C}^{\text{onto}}_t \cup \mathcal{C}^{\text{kg}}_t.
\end{equation}

Ontology concepts and KG relations are constructed semi-automatically using LLM-assisted template extraction and semantic matching from domain policy corpora, requiring limited manual curation primarily for ontology schema definition and validation.

\subsubsection{Constraint-Aligned Retrieval}

For each section, we construct a constraint-aligned retrieval query
\begin{equation}
\label{eq:car}
q_t^\star=\mathrm{ConstructQuery}(x,h_t,\mathcal{C}_t),
\end{equation}
and retrieve evidence as
\begin{equation}
R_t=R(q_t^\star).
\end{equation}

The query construction stage augments the original section query using ontology concepts and KG-linked policy obligations. Ontology concepts and KG-linked obligations are appended as constraint-aligned retrieval terms, guiding retrieval toward structurally relevant evidence. For example, for a \textit{Third-Party Transfer} section, standard RAG typically retrieves general data-sharing passages, while CA-RAG augments retrieval using ontology concepts such as \textit{Transfer Mechanisms} and KG-linked obligations, including \textit{data governance regulations} and \textit{DPIA\footnote{Data Protection Impact Assessment.}} requirements. This guides retrieval toward evidence that is more structurally aligned with downstream compliance requirements. Additional implementation details, retrieval traces, and worked query construction examples are provided in Appendix~\ref{app:query_details}.

\subsection{Section-Wise Generation}

Documents are generated section-wise using shared scenario context and section-specific evidence:
\begin{equation}
\label{eq:section_gen}
y_t=\mathrm{LLM}(x,h_t,R_t,\mathcal{C}_t).
\end{equation}

Each section is generated independently using the shared scenario context, retrieved evidence, and structural constraints. An optional lightweight RLHF refinement stage is applied after generation to improve readability and stylistic consistency.

\paragraph{Pipeline Overview.}
The CA-RAG pipeline can be summarised as:

\smallskip
\noindent
\fbox{
\parbox{\columnwidth}{
\centering
\small
\textsc{
Constraint Extraction
$\rightarrow$
Query Construction
$\rightarrow$
Evidence Retrieval
$\rightarrow$
Section Generation
}
}
}
\smallskip

Figure~\ref{fig:system} illustrates the overall CA-RAG framework. Ontology-derived schema constraints and policy-grounded relational dependencies are extracted before retrieval and incorporated into constraint-aligned retrieval queries. The retrieved evidence is then used for section-wise grounded generation. The full algorithmic specification is provided in Appendix~\ref{app:algorithm}.

\section{Experimental Setup}

\subsection{Domains and Tasks}

We evaluate \textsc{CA-RAG} on Data Sharing Agreement (DSA) synthesis across three compliance-critical domains: \textbf{health}, \textbf{education}, and \textbf{finance}. These domains exhibit varying levels of regulatory complexity, clause dependency structure, and omission sensitivity, enabling analysis of when retrieval-stage structural alignment is most beneficial.  Healthcare emphasises privacy-sensitive compliance, education focuses on governance consistency, and finance contains dense regulatory dependencies. For each domain, we curate corpora of regulatory policies, governance documents, operational guidelines, and domain ontologies. These corpora serve as retrieval sources and inputs for policy-grounded knowledge graph construction.

Clause-level evaluation uses auditing dimensions informed by the clause and
legal-text taxonomies represented in \textsc{CUAD}, \textsc{LEDGAR}, and
\textsc{LexGLUE}; these resources are not used as direct DSA ground-truth
annotations. Mandatory clause completeness is assessed using a GPT-4o-mini clause auditor with deterministic prompting and structured outputs. Clause coverage is additionally measured using checklist-based auditing queries (Appendix~\ref{app:clause_auditor}), enabling interpretable clause-level analysis beyond aggregate similarity metrics. 

\paragraph{Structured Resources.}
The shared ontology contains 21 classes, 15 object properties,
7 datatype properties, and 169 triples. The education, healthcare, and
finance KGs contain 79, 141, and 267 entities, respectively.

\subsection{Models, Baselines, and Generation Protocol}

We evaluate \textsc{CA-RAG} using \textbf{Llama-3.1-8B} \cite{grattafiori2024llama3herdmodels}, \textbf{Llama-3.2-3B}, and \textbf{Mistral-7B} \cite{jiang2023mistral7b}. Dense retrieval uses \texttt{all-MiniLM-L6-v2} for retrieval, ontology matching, and policy subgraph retrieval. We compare standard RAG, ontology-enhanced retrieval (RAG+O), KG-enhanced retrieval (RAG+KG), full CA-RAG, adaptive retrieval baselines (\textsc{Self-RAG}, \textsc{Flare-RAG}), and a generation-only constraint baseline (\textsc{Gen-Constraint}), in which constraints are injected only during generation while retrieval remains identical to standard \textsc{RAG}. Generation proceeds section-wise using shared scenario context, retrieved evidence, and associated structural constraints. For retrieval-conditioned variants, ontology- and policy-grounded constraints are incorporated before retrieval to guide evidence selection. In contrast, \textsc{Gen-Constraint} applies the same constraints only during generation, isolating the effect of retrieval-stage structural alignment.

All methods use the same base models, retriever backbone, decoding settings, prompt scaffold, and evaluation pipeline. An optional lightweight RLHF refinement stage may improve readability without modifying retrieval or evidence support.

\subsection{Evaluation Metrics}
\label{sec:eval_metrics}

We evaluate generated Data Sharing Agreements (DSAs) using metrics that separately capture surface similarity, structural compliance, document-level reliability, and efficiency. Let $\hat{y}$ denote a generated DSA and $y^T$ the corresponding reference agreement.

\paragraph{Similarity Score. (Sim.)}
We measure surface and semantic similarity using cosine similarity and ROUGE-LSUM:
\begin{equation}
\label{eq:sim}
\small
\mathrm{Sim}(\hat{y}, y^T)
=
\tfrac{1}{2}\big(
\mathrm{CosSim}(\hat{y}, y^T)
+
\mathrm{ROUGE\text{-}LSUM}(\hat{y}, y^T)
\big),
\end{equation}
with both components normalised to $[0,1]$.

\paragraph{Compliance Score. (Comp.)}
To assess structural and regulatory completeness, we aggregate five auditing signals: clause compliance from \textsc{CUAD} \cite{hendrycks2021cuad}, provision coverage from \textsc{LEDGAR} \cite{tuggener2020ledgar}, fairness-related signals from \textsc{LexGLUE-UNFAIR-TOS} \cite{Lippi_2019}, domain-specific policy alignment, and explicit DSA clause coverage:
\begin{equation}
\label{eq:comp}
\mathrm{Comp}(\hat{y})
=
\Big(
s_{\text{CUAD}}
\cdot
s_{\text{LED}}
\cdot
s_{\text{Lex}}
\cdot
s_{\text{Dom}}
\cdot
s_{\text{Clause}}
\Big)^{1/5}
\end{equation}
where $s_{\text{LED}}$, $s_{\text{Lex}}$, and $s_{\text{Dom}}$ denote LEDGAR, LexGLUE, and domain-specific compliance scores, respectively.
We use the geometric mean so that omissions in any required dimension lower the overall compliance score.

\paragraph{Reliability Score. (Rel.)}
We define document-level reliability as:
\begin{equation}
\label{eq:rel}
\mathrm{Rel}(\hat{y}, y^T)
=
\lambda\,\mathrm{Comp}(\hat{y})
+
(1-\lambda)\,\mathrm{Sim}(\hat{y}, y^T),
\end{equation}
with $\lambda=0.7$ to prioritise structural and compliance completeness over surface similarity. We report $\mathrm{Rel}$ as a summary metric but treat clause coverage and the presence of missing mandatory clauses as the primary evidence of structural robustness. Method rankings remain stable across $\lambda\in\{0.3,0.5,0.7,0.9\}$; full sensitivity results are reported in the Appendix~\ref{sec:lambda}.

\paragraph{Clause-Level Robustness.}
Because our central claim concerns omission reduction, we report clause coverage, the number of missing mandatory clauses, overall coverage, and dimension-level compliance scores separately.

Implementation details for auditors, compliance scoring, and efficiency measurements, such as latency and empirical throughput, are provided in Appendix~\ref{app:metrics}.

\subsection{Retrieval-Level Evaluation}
\label{sec:retrieval_protocol}

To determine whether improvements arise at retrieval time rather than solely
during generation, we evaluate the retriever independently with generation
disabled. For each scenario--section pair, we pool and deduplicate the union
of the top-20 passages retrieved by standard RAG and \textsc{CA-RAG}. Each
unique passage is manually labelled as \emph{relevant}, \emph{not relevant},
or \emph{uncertain} according to whether it provides evidence supporting the
corresponding scenario--section requirement. Uncertain passages are excluded
from the primary analysis.

Because exhaustive corpus-level relevance judgements are unavailable, we
report Precision@$k$, pooled Recall@$k$, and F1@$k$ for
$k\in\{3,6,9\}$. Pooled Recall@$k$ is computed relative to the set of
passages judged relevant within the pooled top-20 candidate set for each
scenario--section pair rather than against an exhaustively judged corpus.

We additionally report Constraint Coverage, Clause-Evidence Coverage, and
Retrieval Omission Rate to evaluate whether retrieved evidence supports the
structural requirements of downstream generation.

\subsection{Implementation and Controlled Comparison}

All variants use the same retriever backbone, prompt scaffold, decoding
parameters, retrieval corpus, and evaluation pipeline. Dense retrieval uses
\texttt{all-MiniLM-L6-v2}, with $k=6$ passages and up to $m=3$ ontology concepts and $m=3$ KG relations per section. Ground-truth DSAs are excluded
from the retrieval corpus. Experiments run on NVIDIA A100 80GB GPUs.

\section{Results}
\label{sec:results}

We evaluate \textsc{CA-RAG} across education, finance, and healthcare domains using three open-weight large language models and multiple retrieval-generation baselines. Our analysis focuses on whether retrieval-stage structural conditioning reduces omission failures and improves clause-level completeness in long-form constrained generation. In compliance-critical settings, omission failures are often more consequential than small reductions in semantic similarity. Consequently, our evaluation emphasises structural completeness, clause recovery, and omission robustness in addition to aggregate reliability metrics. Unless otherwise stated, results are macro-averaged within each domain. Full per-domain results are provided in the Appendix~\ref{app:full_results}.

\subsection{Retrieval-Level Evaluation}
\label{sec:retrieval_eval}

We first evaluate retrieval independently of downstream generation.
Table~\ref{tab:retrieval_ir} reports manually verified retrieval performance
aggregated across domains. \textsc{CA-RAG} improves Precision@$k$, pooled
Recall@$k$, and F1@$k$ at every evaluated retrieval depth. At the generation setting ($k=6$), Precision increases from $0.5268$ to $0.5692$, pooled Recall from $0.5402$ to $0.5950$, and F1 from $0.5389$ to $0.5672$. Because generation is disabled in this experiment, these gains cannot be attributed to prompting or decoding and provide direct evidence that structural conditioning changes evidence selection before generation.

\begin{table}[t]
\centering
\footnotesize
\caption{Retrieval-only evaluation aggregated across domains using manually
annotated pooled evidence. Pooled Recall is computed relative to relevant
passages within the judged candidate pool. Higher is better.}
\label{tab:retrieval_ir}
\setlength{\tabcolsep}{3pt}
\begin{tabular}{lcccc}
\toprule
\textbf{Method} & $\mathbf{k}$ &
\textbf{P@$k$} & \textbf{Pooled R@$k$} & \textbf{F1@$k$} \\
\midrule
RAG     & 3 & 0.5000 & 0.2569 & 0.3966 \\
\textsc{CA-RAG} & 3 & \textbf{0.5311} & \textbf{0.2754} & \textbf{0.4193} \\
\midrule
RAG     & 6 & 0.5268 & 0.5402 & 0.5389 \\
\textsc{CA-RAG} & 6 & \textbf{0.5692} & \textbf{0.5950} & \textbf{0.5672} \\
\midrule
RAG     & 9 & 0.5499 & 0.8654 & 0.6506 \\
\textsc{CA-RAG} & 9 & \textbf{0.5866} & \textbf{0.9185} & \textbf{0.6938} \\
\bottomrule
\end{tabular}
\end{table}

\begin{figure*}[t]
    \centering
    \includegraphics[width=0.80\linewidth]{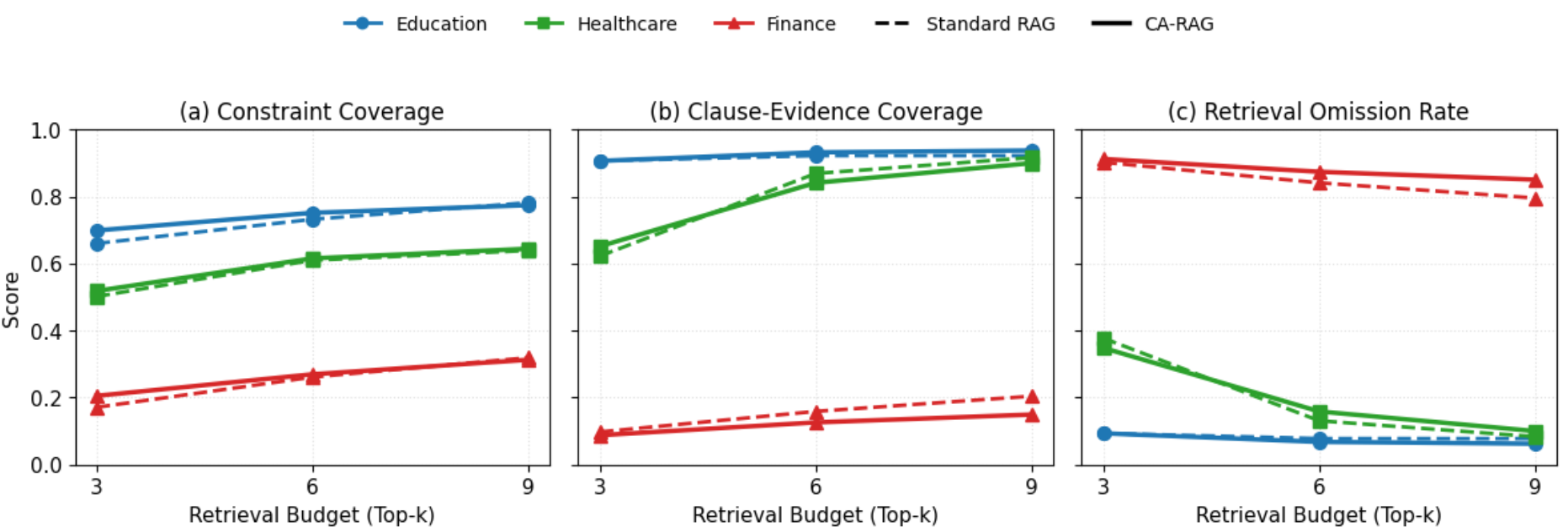}
    \caption{Structural sufficiency across domains and retrieval budgets
    ($k\in\{3,6,9\}$), measured by Constraint Coverage, Clause-Evidence
    Coverage, and Retrieval Omission Rate. Solid lines denote
    \textsc{CA-RAG}; dashed lines denote RAG.}
    \label{fig:retrieval_metrics}
\end{figure*}

\subsection{Structural Sufficiency of Retrieved Evidence}
\label{sec:structural_retrieval}

Standard relevance metrics indicate whether retrieved passages are relevant,
but not whether they sufficiently support mandatory downstream requirements.
We therefore evaluate \emph{Constraint Coverage} (CCov) and
\emph{Clause-Evidence Coverage} (CECov), and report
\emph{Retrieval Omission Rate} (ROR) as the complementary measure of missing
clause-level evidence. CCov measures the proportion of required ontology/KG
constraints supported by retrieved evidence, while CECov measures the
proportion of mandatory clause requirements with supporting evidence.

For required constraints $\mathcal{C}$ and top-$k$ passages $R_k$,
\begin{equation}
\mathrm{CCov}@k =
\frac{
|\{c\in\mathcal{C}:\exists r\in R_k,\,
\mathrm{Supports}(r,c)\}|
}{
|\mathcal{C}|
}.
\end{equation}

For mandatory clauses $\mathcal{Q}$,
\begin{equation}
\mathrm{CECov}@k =
\frac{
|\{q\in\mathcal{Q}:\exists r\in R_k,\,
\mathrm{Supports}(r,q)\}|
}{
|\mathcal{Q}|
}.
\end{equation}

Retrieval Omission Rate is:
\begin{equation}
\mathrm{ROR}@k = 1-\mathrm{CECov}@k.
\end{equation}

Support is determined from manual evidence annotations indicating whether a
retrieved passage directly supports the corresponding constraint or mandatory
clause. Together, these metrics distinguish relevant retrieval from evidence
that is structurally sufficient for downstream generation.

Figure~\ref{fig:retrieval_metrics} shows substantial domain variation.
Education maintains high CECov and low omission under both methods, with
\textsc{CA-RAG} providing modest CCov gains. In healthcare,
\textsc{CA-RAG} improves CECov and reduces omissions, particularly at larger
retrieval budgets. Finance remains more challenging: CCov is comparable to
RAG, but \textsc{CA-RAG} retrieves less direct clause-supporting evidence.
Thus, satisfying ontology- and KG-derived constraints does not guarantee
coverage of every mandatory clause. Overall, constraint alignment can improve
the structural utility of retrieved evidence, but its effectiveness depends
on alignment between structural constraints and available domain evidence.
This also explains why improvements in conventional retrieval metrics do not
translate uniformly into higher clause-evidence coverage.

\subsection{Overall Generation Performance}
\label{sec:overall}
Table~\ref{tab:accuracy_grid} reports document-level reliability across models and domains. \textsc{CA-RAG} achieves the highest mean reliability in six of nine
model--domain configurations, including all three finance configurations,
while competing methods remain strongest in the remaining healthcare and
education settings. These results should be interpreted together with Sections~\ref{sec:retrieval_eval}--\ref{sec:structural_retrieval}: retrieval conditioning measurably changes evidence selection before generation, while
the magnitude of downstream gains remains model and domain dependent. Finance domain provides a particularly informative setting because of its dense
regulatory dependencies and omission-sensitive requirements. Although
retrieval remains challenging in this domain, \textsc{CA-RAG} yields strong
downstream reliability in several model configurations.

Adaptive retrieval methods such as \textsc{Self-RAG} and \textsc{Flare-RAG} improve aggregate reliability in some domains, particularly health, but also exhibit higher variance and latency due to iterative retrieval and refinement. In contrast, \textsc{CA-RAG} maintains comparatively stable performance while reducing omission frequency with lower retrieval overhead. Overall, the primary contribution of \textsc{CA-RAG} lies not in maximising aggregate generation scores, but in explicitly aligning evidence selection with downstream structural requirements, which can improve structural completeness and reduce omission failures.

We additionally performed paired bootstrap significance testing across domains, runs, and model variants using document-level reliability scores. CA-RAG significantly outperformed standard RAG, \textsc{Self-RAG}, and \textsc{Flare-RAG}, while differences relative to the generation-only constraint baseline were not statistically significant. This suggests that generation-time prompting can partially improve aggregate reliability, whereas retrieval-stage conditioning directly modifies evidence selection and can improve omission robustness and structural completeness. (Section~\ref{sec:ablation}). Full paired bootstrap significance results are provided in Appendix~\ref{app:significance}.

\begin{table*}[t]
\centering
\small
\caption{Reliability (mean $\pm$ std over 3 runs). Best results are
\textbf{bold}; second-best are \underline{underlined}. }
\label{tab:accuracy_grid}
\setlength{\tabcolsep}{2pt}
\renewcommand{\arraystretch}{1.0}
\resizebox{\textwidth}{!}{
\begin{tabular}{lccc|ccc|ccc}
\toprule
 & \multicolumn{3}{c|}{\textbf{Llama-3.2-3B}} 
 & \multicolumn{3}{c|}{\textbf{Llama-3.1-8B}} 
 & \multicolumn{3}{c}{\textbf{Mistral-7B}} \\
\cmidrule(lr){2-4} \cmidrule(lr){5-7} \cmidrule(lr){8-10}
\textbf{Method} 
 & Finance & Healthcare & Education
 & Finance & Healthcare & Education
 & Finance & Healthcare & Education \\
\midrule
RAG        
& 0.6156 $\pm$ 0.015 & \underline{0.6634 $\pm$ 0.012} & \underline{0.6515 $\pm$ 0.015}
& \underline{0.6259 $\pm$ 0.018} & 0.6560 $\pm$ 0.034 & \underline{0.6685 $\pm$ 0.023}
& 0.6051 $\pm$ 0.016 & 0.6682 $\pm$ 0.017 & 0.6588 $\pm$ 0.010 \\
\addlinespace[2pt]
\textsc{Self-RAG}   
& \underline{0.6227 $\pm$ 0.023} & 0.6602 $\pm$ 0.010 & 0.6167 $\pm$ 0.119
& 0.6024 $\pm$ 0.023 & \underline{0.6666 $\pm$ 0.021} & 0.6772$\pm$ 0.025 
& 0.6106 $\pm$ 0.050 & 0.6625 $\pm$ 0.019 & \textbf{0.6673 $\pm$ 0.004} \\
\addlinespace[2pt]
\textsc{Flare-RAG}  
& 0.6092 $\pm$ 0.044 & 0.6604 $\pm$ 0.020 & 0.6459 $\pm$ 0.016 
& 0.6145 $\pm$ 0.034 & \textbf{0.6793 $\pm$ 0.025} & 0.6626 $\pm$ 0.027
& \underline{0.6113 $\pm$ 0.034} & \textbf{0.6780 $\pm$ 0.020} & 0.6504 $\pm$ 0.052\\
\addlinespace[2pt]
\textsc{CA-RAG$^{*}$}   
& \textbf{0.6410 $\pm$ 0.006}& \textbf{0.6730 $\pm$ 0.012} & \textbf{0.6521 $\pm$ 0.022} 
& \textbf{0.6404 $\pm$ 0.008} & 0.6549 $\pm$ 0.016 & \textbf{0.6830 $\pm$ 0.006}
& \textbf{0.6329 $\pm$ 0.021} & \underline{0.6698 $\pm$ 0.008} & \underline{0.6653 $\pm$ 0.013}\\
\bottomrule
\end{tabular}
}
\end{table*}

\subsection{Structural and Compliance Trade-off Analysis}
\label{sec:tradeoff}

We analyse retrieval conditioning across six auditing metrics: clause compliance, provision coverage, fairness and balance, domain alignment, DSA clause coverage, and overall compliance. Figure~\ref{fig:l31r} shows results for \textbf{Llama-3.1-8B}; additional plots are provided in the appendix. The radar plots reveal complementary behaviour between ontology-derived and policy-grounded constraints. Ontology constraints primarily improve provision coverage and structural completeness, while KG-based constraints strengthen policy-specific obligations and clause-level compliance behaviour. The full \textsc{CA-RAG} configuration generally provides a balanced structural profile across domains. Gains are strongest in finance and education, where omission-sensitive clauses depend on structurally connected evidence across multiple sections. Overall, the analysis highlights the importance of multi-dimensional auditing beyond aggregate similarity metrics.

\begin{figure*}[t]
  \centering
  \includegraphics[width=0.8\linewidth]{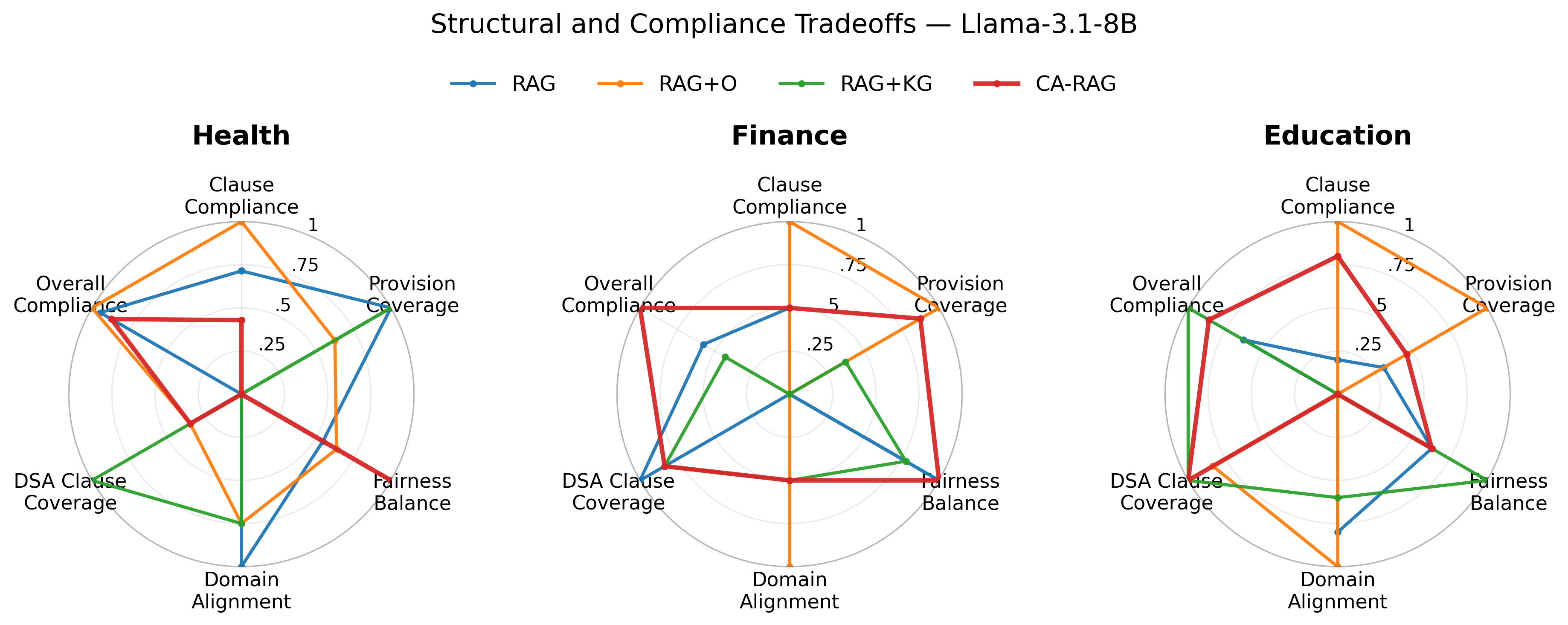}
  \caption{Structural and compliance trade-offs across retrieval variants for \textbf{Llama-3.1-8B}. Scores are normalised across six auditing metrics. Ontology constraints improve structural completeness, while KG constraints strengthen clause compliance.}
  \label{fig:l31r}
\end{figure*}

\subsection{Ablation: Retrieval-Stage vs Generation-Only Constraints}
\label{sec:ablation}

Generation-only constraints achieve the highest clause coverage
($78.7\%$), showing that explicit structural prompts can encourage clause
inclusion during decoding. However, \textsc{CA-RAG} achieves higher
compliance ($0.6726$) and reliability ($0.6569$). The retrieval-only results provide evidence for a mechanism that generation-only prompting cannot provide: \textsc{CA-RAG} changes the evidence available before decoding. At $k=6$, it improves both Precision@$k$ and pooled Recall@$k$, while structural retrieval analysis shows improved availability of clause-supporting evidence. Thus, generation-only and
retrieval-stage constraints affect different parts of the pipeline.

\begin{table}[t]
\centering
\footnotesize
\setlength{\tabcolsep}{4pt}
\caption{Ablation comparing generation-only constraints and retrieval-stage structural alignment. Higher is better.}
\label{tab:ablation}

\begin{tabular}{lccc}
\toprule
\textbf{Variant} 
& \textbf{Clause Cov.} 
& \textbf{Comp.} 
& \textbf{Rel.} \\
\midrule

\textsc{RAG} 
& 75.9 & 0.6571 & 0.6459 \\

\textsc{Gen-Constraint} 
& \textbf{78.7} & 0.6683 & 0.6517 \\

\textsc{CA-RAG} 
& \underline{77.8} 
& \textbf{0.6726} 
& \textbf{0.6569} \\
\bottomrule
\end{tabular}
\end{table}

\subsection{Ablation: Retrieval-Budget Sensitivity}

On finance with Llama-3.1-8B, standard RAG shows a small compliance gain from
$k=3$ to $k=6$ but degrades substantially at $k=9$, whereas \textsc{CA-RAG} remains comparatively stable. Its best reliability occurs at $k=6$ ($0.6404$), indicating that the gains are not explained simply by retrieving more evidence. Full results are reported in the appendix \ref{sec:topk}.

\subsection{Efficiency and Auditor Validation}
\label{sec:efficiency}

\textsc{CA-RAG} adds constraint extraction and query-construction overhead but
avoids the iterative retrieval loops used by \textsc{Self-RAG} and
\textsc{Flare-RAG}. Its latency remains closer to standard RAG than to
adaptive retrieval baselines in most settings; full efficiency results are
reported in the Appendix~\ref{app:efficiency}.

We independently validate automated clause-coverage estimates against
manual assessment. The auditor obtains MAE $=2.63$ percentage points,
Pearson $r=0.967$ ($p<0.001$), and Spearman $\rho=0.777$ ($p=0.003$),
indicating close agreement with manual evaluation.

\section{Conclusion}

We presented \textsc{CA-RAG}, a retrieval-conditioning framework that aligns
evidence selection with structural and policy requirements for
compliance-sensitive long-form generation. Retrieval-only evaluation
with manually annotated evidence shows improved evidence selection across
$k\in\{3,6,9\}$, with \textsc{CA-RAG} improving Precision@$k$, pooled
Recall@$k$, and F1@$k$ over standard RAG. Structural evidence sufficiency
varies across domains, with gains in constraint and clause-evidence coverage
most apparent in education and healthcare, while finance remains more
challenging. Generation experiments show corresponding improvements in compliance and reliability in several model--domain settings, although gains remain model and domain dependent. These findings suggest that semantic relevance alone can be insufficient for omission-sensitive long-form generation, and that
retrieval-stage structural alignment can improve evidence sufficiency before
generation.

\section*{Limitations}

While \textsc{CA-RAG} improves structural completeness and reduces omission failures, several limitations remain.

\noindent\textbf{Dependence on structured resources.}
Our approach relies on the availability and quality of domain ontologies and policy-grounded knowledge graphs. In domains where such structured resources are incomplete, noisy, or unavailable, retrieval conditioning may be less effective. Although ontology and KG construction are largely automated, missing or incorrect relations can propagate into retrieval conditioning.

\noindent\textbf{No guarantee of complete retrieval.}
Constraint-aligned retrieval increases the likelihood of retrieving structurally relevant evidence, but does not guarantee recovery of all required clauses. Even with structural conditioning, omission-sensitive obligations may still be missed when supporting evidence is incomplete or weakly represented in the corpus.

\noindent\textbf{Scope of evaluation.}
Our experiments focus on Data Sharing Agreement synthesis across healthcare, finance, and education domains using three open-weight LLM families. While these domains exhibit diverse structural and regulatory characteristics, the findings may not directly generalise to other long-form generation tasks, languages, model scales, or retrieval settings without further evaluation.

\noindent\textbf{Potential evaluation limitations.}
Clause-level compliance evaluation relies in part on LLM-based auditing pipelines and benchmark-derived compliance signals. Although the same deterministic evaluation protocol is applied consistently across methods, automatic auditing may still miss subtle legal or domain-specific nuances. In addition, aggregate similarity metrics do not always correlate with structural completeness or omission robustness.

\noindent\textbf{Computational overhead and retrieval scaling.}
Ontology matching, KG retrieval, and constraint-aligned query construction introduce additional latency relative to standard RAG pipelines. Retrieval-budget sensitivity experiments further show that increasing retrieval depth does not necessarily improve downstream performance, particularly for standard RAG. Scalability to larger corpora and real-time deployment settings requires further study.

\noindent\textbf{Generality of refinement methods.}
Our RLHF experiments suggest that post-generation refinement mainly improves stylistic coherence and local fluency rather than reliably recovering missing evidence-dependent clauses. More advanced refinement or agentic verification mechanisms may therefore be necessary for robust omission-sensitive generation.

Overall, our findings suggest that retrieval-stage structural alignment is a promising direction for improving robustness in long-form generation settings where omission failures and incomplete evidence can lead to unreliable outputs. Future work includes extending retrieval-stage structural conditioning to broader settings, including compliance, policy generation, and agentic document generation.

\section*{Ethical Considerations}

Constraint-aligned retrieval may improve the reliability of long-form compliance and policy generation systems by reducing omission-sensitive failures. However, generated agreements may still contain incomplete, outdated, or legally insufficient content if supporting evidence is missing or retrieval fails. Such systems should therefore be used as assistive drafting tools rather than substitutes for expert legal or regulatory review. In addition, policy-grounded generation systems could be misused to automate large-scale production of synthetic policies or compliance documents without appropriate human oversight.

\bibliography{custom}

\appendix

\section{Appendix}
\label{sec:appendix}

\subsection{Constraint-Aligned RAG Pipeline}
\label{app:algorithm}

Algorithm~\ref{alg:carag} summarises the full \textsc{CA-RAG} retrieval-conditioning pipeline used in all experiments. Ontology concepts and KG relations are automatically constructed via template-driven extraction and semantic matching from domain policy corpora, with minimal manual annotation beyond ontology schema definition.

\begin{algorithm}[t]
\footnotesize
\caption{\textsc{CA-RAG}: Constraint-Aligned Retrieval-Augmented Generation}
\label{alg:carag}
\begin{algorithmic}[1]

\Require Scenario $x$; corpus $\mathcal{R}$; ontology $\mathcal{O}$;
policy corpus $\mathcal{P}$; retrieval encoder $E_{\text{ret}}$;
ontology encoder $E_{\text{onto}}$; LLM $p_\theta$; variant $\nu$;
top-$k$, top-$m$
\Ensure Generated document $\hat{y}$

\State $\hat{y} \gets \emptyset$
\State $\mathcal{V}_{\text{onto}}
       \gets \{\textsc{RAG+O},\textsc{CA-RAG}\}$
\State $\mathcal{V}_{\text{kg}}
       \gets \{\textsc{RAG+KG},\textsc{CA-RAG}\}$

\Statex \textbf{Step 1: Section Planning}
\State $H \gets \mathrm{GenerateSectionsTitles}(p_\theta,x)$

\Statex \textbf{Step 2: Policy Graph Initialisation}
\If{$\nu \in \mathcal{V}_{\text{kg}}$}
    \State $G \gets \mathrm{BuildPolicyGraph}(\mathcal{P})$
\EndIf

\For{$t=1$ to $|H|$}
    \State $h_t \gets H[t]$
    \State $\mathcal{C}_t \gets \emptyset$

    \Statex \textbf{Step 3: Ontology Constraint Extraction}
    \If{$\nu \in \mathcal{V}_{\text{onto}}$}
        \State $\mathcal{C}^{\text{onto}}_t
        \gets \mathrm{OntologyMatch}
        (E_{\text{onto}},h_t,\mathcal{O},m)$
        \State $\mathcal{C}_t
        \gets \mathcal{C}_t\cup\mathcal{C}^{\text{onto}}_t$
    \EndIf

    \Statex \textbf{Step 4: KG Constraint Retrieval}
    \If{$\nu \in \mathcal{V}_{\text{kg}}$}
        \State $\mathcal{C}^{\text{kg}}_t
        \gets \mathrm{RetrievePolicyRelations}(G,x,h_t,m)$
        \State $\mathcal{C}_t
        \gets \mathcal{C}_t\cup\mathcal{C}^{\text{kg}}_t$
    \EndIf

    \Statex \textbf{Step 5: Constraint-Aligned Query Construction}
    \State $q_t^\star
        \gets \mathrm{ConstructQuery}(x,h_t,\mathcal{C}_t)$
    \Statex \hspace{\algorithmicindent}
        \textit{Append ontology concepts and KG-linked obligations.}

    \Statex \textbf{Step 6: Evidence Retrieval}
    \State $R_t
        \gets \mathrm{DenseRetrieve}
        (E_{\text{ret}},q_t^\star,\mathcal{R},k)$

    \Statex \textbf{Step 7: Section Generation}
    \State $y_t
        \gets \mathrm{GenerateSection}
        (p_\theta,x,h_t,R_t,\mathcal{C}_t)$
    \State $\hat{y}
        \gets \hat{y}\Vert\mathrm{FormatSection}(h_t,y_t)$
\EndFor

\State \Return $\hat{y}$

\end{algorithmic}
\end{algorithm}

\subsection{Extended Query Construction Details}
\label{app:query_details}

This section provides implementation details for the \textsc{ConstructQuery} stage. \textsc{CA-RAG} combines ontology-derived schema concepts with policy-grounded KG relations before retrieval, thereby conditioning which evidence becomes retrievable for section-wise generation.

\subsubsection{Ontology Concept Injection}

Ontology concepts are extracted from the constructed ontology using semantic matching over OWL classes and properties. For each section heading $h_t$, ontology labels, local names, and optional comments are embedded using \texttt{all-MiniLM-L6-v2}. The implementation combines embedding similarity with lexical similarity:
\begin{equation}
\mathrm{Score}(h_t,o_i)
=
0.75\,s_{\mathrm{emb}}(h_t,o_i)
+
0.25\,s_{\mathrm{lex}}(h_t,o_i),
\end{equation}
where $s_{\mathrm{emb}}$ is cosine similarity and $s_{\mathrm{lex}}$ is Levenshtein similarity. The top-$m$ ontology concepts are retained and appended as schema-level retrieval terms.

For example, the ontology contains concepts such as \textit{Data Recipient}, \textit{Data Transfer Event}, \textit{Data Flow}, \textit{Audit Trail}, \textit{Risk Assessment}, and \textit{Data Minimisation Rule}, which provide structural cues for section-specific retrieval.

\subsubsection{KG Relation Expansion}

Policy-grounded KG expansion retrieves relevant RDF triples from the domain KG. Each triple $\tau_i=(s_i,p_i,o_i)$ is scored against the section heading by comparing the heading embedding with the subject, predicate, and object embeddings:
\begin{equation}
\mathrm{Score}(h_t,\tau_i)
=
\max
\big(
s(h_t,s_i),
s(h_t,p_i),
s(h_t,o_i)
\big).
\end{equation}

Triples above a similarity threshold are retained; if no triple satisfies the threshold, the highest-ranked triples are returned to avoid empty constraint contexts. This allows the retriever to recover policy obligations that may not be surfaced by semantic similarity alone.

\subsubsection{Constraint-aligned Query Construction}

The final retrieval query is constructed by concatenating the scenario, section heading, ontology concepts, and KG-linked obligations:
\begin{equation}
q_t^\star
=
[x;\,h_t;\,\mathcal{C}^{\mathrm{onto}}_t;\,\mathcal{C}^{\mathrm{kg}}_t].
\end{equation}

Duplicate or low-similarity constraints are filtered before dense retrieval. In practice, this yields a constraint-aligned query of the original section query with structural and policy-specific terms.

\subsubsection{Worked Example}

Consider a finance DSA section on \textit{Third-Party Transfer} or recipient obligations.

\paragraph{Standard RAG query}
\begin{quote}
``Generate a third-party transfer policy for financial data sharing.''
\end{quote}

This may retrieve general passages about data sharing or external access.

\paragraph{CA-RAG expansion}
Ontology matching may add:
\begin{quote}
\texttt{Data Recipient}, \texttt{Data Transfer Event}, \texttt{Data Flow}, \texttt{Audit Trail}
\end{quote}

KG retrieval may add obligations such as:
\begin{quote}
\small
\begin{itemize}
    \item (\textit{Recipient Agency}, \textsc{prohibits}, \textit{Onward Sharing Without Legal Basis})
    
    \item (\textit{Recipient Agency}, \textsc{requires}, \textit{Incident Response Procedures})
    
    \item (\textit{Recipient Agency}, \textsc{requires}, \textit{Retention Period Definition})
    
    \item (\textit{Recipient Agency}, \textsc{supports}, \textit{Data Subject Rights})
\end{itemize}
\end{quote}
The resulting pseudo-query becomes:
\begin{quote}
\texttt{Third-party transfer policy for financial data sharing + Data Recipient + Data Transfer Event + Audit Trail + onward sharing restrictions + incident response procedures + retention periods + data subject rights.}
\end{quote}

This shifts retrieval from generic data-sharing evidence to policy passages that contain transfer restrictions, recipient obligations, incident response duties, and retention requirements.

\subsubsection{Retrieval Trace Example}

Table~\ref{tab:retrieval_trace} illustrates a representative retrieval trace for finance-domain transfer obligations. Standard \textsc{RAG} retrieves semantically related passages describing generic data sharing and recipient access policies, but frequently omits structurally dependent obligations required for downstream compliance generation.

In contrast, \textsc{CA-RAG} expands retrieval by incorporating ontology-derived schema concepts and KG-linked policy relations before evidence selection. This shifts retrieval toward passages containing transfer restrictions, incident response duties, retention requirements, and data-subject-rights obligations. The resulting evidence is therefore more structurally aligned with downstream clause requirements, reducing omission failures during section generation.

\begin{table}[t]
\centering
\footnotesize
\caption{Representative retrieval trace for finance-domain recipient transfer obligations.}
\label{tab:retrieval_trace}
\setlength{\tabcolsep}{3pt}
\renewcommand{\arraystretch}{1.05}
\begin{tabular}{l p{4.6cm}}
\toprule
\textbf{Method} & \textbf{Retrieved Evidence} \\
\midrule
RAG &
General data-sharing and recipient access passages. \\
CA-RAG &
Recipient onward-sharing restrictions, incident response duties, retention-period obligations, and data-subject-rights support. \\
\bottomrule
\end{tabular}
\end{table}

\subsection{Additional Worked Retrieval Examples}
\label{app:worked_examples}

This section provides qualitative examples illustrating how retrieval-stage structural conditioning affects omission-sensitive clause recovery across finance, healthcare, and education DSAs.

Across domains, standard \textsc{RAG} frequently retrieves semantically related but structurally incomplete evidence, leading to repeated omission of globally dependent obligations such as cross-border transfer safeguards, DPIA/TIA requirements, incident response procedures, and legal basis clauses. In contrast, \textsc{CA-RAG} substantially reduces omission frequency by retrieving evidence aligned with downstream structural dependencies.

\begin{table*}[t]
\centering
\footnotesize
\caption{Representative omission patterns across domains.}
\label{tab:worked_examples}
\setlength{\tabcolsep}{4pt}
\renewcommand{\arraystretch}{1.08}

\begin{tabular}{p{1.5cm} p{1.4cm} p{4.5cm} p{4.8cm}}
\toprule
\textbf{Domain} &
\textbf{Method} &
\textbf{Retrieved Evidence} &
\textbf{Representative Missing Clauses} \\
\midrule

Finance
&
RAG
&
General transfer policies, recipient access rules, and generic governance passages
&
\textit{Cross Border Transfers},
\textit{DPIA/TIA},
\textit{Data Minimisation},
\textit{Legal Basis},
\textit{Onward Disclosure},
\textit{Subprocessors Third Parties} \\

\midrule

Finance
&
CA-RAG
&
Transfer safeguards, retention obligations, governance dependencies, and audit requirements
&
Primarily residual administrative clauses such as
\textit{Data Format Schema},
\textit{Attribution Publication},
\textit{Intellectual Property} \\

\midrule

Healthcare
&
RAG
&
Generic patient-data handling and storage policies
&
\textit{Incident Response}, \textit{Breach Notification}, \textit{Data Subject Rights}, \textit{Transparency Notices}, \textit{Cross Border Transfers} \\

\midrule

Healthcare
&
CA-RAG
&
Incident response duties, breach reporting, governance and audit procedures
&
Reduced omission frequency, with remaining omissions concentrated in secondary administrative clauses \\

\midrule

Education
&
RAG
&
Institutional data-access and student-data usage policies
&
\textit{Incident Response},
\textit{Cross Border Transfers},
\textit{Legal Basis},
\textit{Data Subject Rights},
\textit{Subprocessors Third Parties} \\

\midrule

Education
&
CA-RAG
&
Governance procedures, retention obligations, transfer safeguards, review requirements
&
Primarily procedural and publication-related clauses, such as
\textit{Attribution Publication}
and
\textit{Sanctions Remediation} \\

\bottomrule
\end{tabular}
\end{table*}

The examples in Table~\ref{tab:worked_examples} illustrate that the primary advantage of \textsc{CA-RAG} is not simply retrieving more passages, but retrieving evidence structurally aligned with downstream compliance requirements. Standard relevance-based retrieval repeatedly omits globally dependent obligations such as DPIA/TIA requirements, incident-response duties, legal basis clauses, and cross-border transfer safeguards. Retrieval-stage structural conditioning substantially improves recovery of these omission-sensitive clauses before generation begins.

\subsection{RLHF Refinement Implementation}
\label{app:rlhf_impl}

We implement an optional RLHF refinement module applied after section-wise generation. Domain experts provide scalar quality ratings in $[0,1]$ for generated sections, which are used to train a lightweight reward model and perform PPO-style policy refinement while keeping retrieval and constraint injection fixed.

\textbf{Reward model.}
Given prompt--section pairs $(p_t,\hat{y}_t)$ and human ratings $r_t$, the reward model predicts a scalar quality score $\hat{r}_t$ and is trained using mean squared error:
\begin{equation}
\label{eq:rm_loss}
\mathcal{L}_{\textsc{RM}}
=
\frac{1}{N}
\sum_{t=1}^{N}
(\hat{r}_t-r_t)^2.
\end{equation}

\textbf{Policy refinement.}
We generate multiple candidate rewrites per section using stochastic decoding and score them using the reward model. PPO-style optimisation is then applied using a clipped objective with KL regularisation:
\begin{equation}
\label{eq:ppo_style}
\begin{aligned}
\mathcal{L}_{\textsc{PPO}}
=&\;
-\min\Big(
\rho A,\;
\mathrm{clip}(\rho,1-\epsilon,1+\epsilon)A
\Big)
\\
&+
\beta\,\mathrm{KL}
\big(
\pi_{\theta}\|\pi_{\text{ref}}
\big).
\end{aligned}
\end{equation}
where $\rho$ denotes the policy ratio, $A$ the reward-based advantage signal, $\epsilon$ the clipping threshold, and $\beta$ the KL coefficient.

This refinement stage is intended as a lightweight stylistic improvement module rather than a replacement for retrieval-conditioned evidence selection.

\subsection{Evaluation: Auditors, Metrics, and Implementation}
\label{app:metrics}

All documents are evaluated on extracted text. PDFs are segmented into overlapping chunks (1000 tokens, overlap 100) for retrieval and downstream evaluation.

\subsubsection{Reference-Based Similarity Metrics}
\label{app:similarity}

We compute document-level similarity between generated DSAs $\hat{y}$ and reference agreements $y^T$ using cosine similarity over sentence embeddings and ROUGE-LSUM. Cosine similarity measures semantic alignment, while ROUGE-LSUM captures lexical overlap.

\subsubsection{Compliance Auditing Signals}
\label{app:compliance_auditors}

We compute five compliance-oriented signals, each normalised to $[0,1]$:
\begin{enumerate}[itemsep=1pt, topsep=1pt]
    \item CUAD-based clause compliance \cite{hendrycks2021cuad},
    \item LEDGAR \cite{tuggener2020ledgar} provision coverage,
    \item LexGLUE UNFAIR\_TOS \cite{Lippi_2019} fairness/balance signals,
    \item Domain alignment against in-domain policy PDFs, and
    \item DSA clause coverage via a checklist-driven auditor.
\end{enumerate}

All datasets and benchmark resources used in this work (e.g., CUAD, LEDGAR, LexGLUE) are publicly available research datasets used in accordance with their intended academic and research usage conditions.

\paragraph{DSA Clause Auditor}
\label{app:clause_auditor}

We measure clause completeness using a checklist-based DSA clause auditor. Given a document split into text chunks $\mathcal{X}=\{x_i\}_{i=1}^{n}$, we define a fixed set of clause questions $\mathcal{Q}=\{(l_j,q_j)\}_{j=1}^{m}$, where $l_j$ is a clause label and $q_j$ is a natural-language question.

For each question $q_j$, we retrieve the top-$k$ relevant chunks using TF--IDF retrieval, concatenate them into a context window, and query an LLM-based judge that returns a structured decision containing an \texttt{answered} boolean and supporting evidence spans. This TF--IDF step is used only for selecting chunks within the evaluation auditor and is independent of the dense retriever evaluated in
\textsc{CA-RAG}.

Clause coverage is computed as:
\begin{equation}
\label{eq:clause_coverage}
\mathrm{ClauseCov}(\hat{y})
=
\frac{1}{m}
\sum_{j=1}^{m}
\mathbb{I}\big[\texttt{answered}(q_j,\hat{y})=1\big].
\end{equation}

The auditor additionally stores per-question evidence and rationale for downstream error analysis.

\subsubsection{Aggregate Scores}
\label{app:aggregate_scores}

We report similarity, compliance, reliability, and efficiency metrics as macro-averaged across all domains.

\subsubsection{Efficiency Measurement}
\label{app:efficiency_eval}

We report average token usage, end-to-end latency, and empirical throughput (effective TFLOPs/s). Latency is measured per section-generation call and includes prompt construction, constraint extraction, retrieval, and decoding:

\begin{equation}
\label{eq:latency}
\begin{aligned}
T_{\text{latency}}
=
\frac{1}{N}
\sum_{n=1}^{N}
\sum_{s=1}^{S}
\Big(
&T_{\text{prompt}}^{(n,s)}
+
T_{\text{constraint}}^{(n,s)}
\\
&+
T_{\text{retrieval}}^{(n,s)}
+
T_{\text{decode}}^{(n,s)}
\Big)
\end{aligned}
\end{equation}

Effective TFLOPs/s is reported as an empirical proxy for throughput under a fixed hardware configuration.

\subsection{Sensitivity to Reliability Weighting}
\label{sec:lambda}

Table~\ref{tab:lambda_ablation} analyses sensitivity to the weighting parameter $\lambda$. Reliability increases with larger $\lambda$ values in finance and education, indicating stronger dependence on compliance-oriented metrics. Healthcare remains comparatively stable across weighting values. We use $\lambda=0.7$ to prioritise structural completeness while retaining sensitivity to semantic similarity. Importantly, method rankings remain stable across the tested range, indicating that the gains of \textsc{CA-RAG} are not dependent on a specific weighting choice.

\begin{table}[t]
\centering
\caption{Sensitivity analysis of the reliability score with different values of $\lambda$. Results are averaged across all evaluation samples for each domain.}
\label{tab:lambda_ablation}
\resizebox{\columnwidth}{!}{
\begin{tabular}{lcccccc}
\toprule
Domain & $\lambda=0.3$ & $\lambda=0.5$ & $\lambda=0.7$ & $\lambda=0.9$ & Comp. & Sim. \\
\midrule
Healthcare & 0.6617 & 0.6609 & 0.6600 & 0.6592 & 0.6589 & 0.6628 \\
Finance & 0.5808 & 0.6002 & 0.6195 & 0.6389 & 0.6486 & 0.5518 \\
Education & 0.6499 & 0.6557 & 0.6615 & 0.6672 & 0.6701 & 0.6413 \\
\bottomrule
\end{tabular}
}
\end{table}

\subsection{Ablation: Retrieval Budget Sensitivity}
\label{sec:topk}

The effect of retrieval budget differs across methods. For standard
\textsc{RAG}, increasing $k$ from 3 to 6 slightly improves compliance
($0.6547\rightarrow0.6615$), while reliability remains approximately stable
($0.6282\rightarrow0.6259$). At $k=9$, both compliance and reliability
decrease substantially. This indicates that increasing retrieval depth alone
does not guarantee better downstream performance; additional retrieved
evidence may remain weakly aligned with the structural requirements of the
generated document.

In contrast, \textsc{CA-RAG} maintains comparatively stable performance across
retrieval budgets while consistently achieving higher compliance and
reliability than standard \textsc{RAG}. The strongest performance occurs at
$k=6$, where \textsc{CA-RAG} achieves a reliability score of $0.6404$.
Even at $k=9$, \textsc{CA-RAG} preserves higher compliance and reliability
than standard \textsc{RAG}. These findings suggest that the benefits of
\textsc{CA-RAG} arise from structurally conditioned evidence selection rather
than increased retrieval quantity alone. In omission-sensitive domains such
as finance, retrieving additional evidence is insufficient when the retrieved
context remains structurally incomplete or weakly aligned with downstream
clause requirements.

\begin{table}[t]
\centering
\small
\caption{Top-k retrieval sensitivity analysis on the finance domain.}
\label{tab:topk}
\begin{tabular}{l c c c c}
\toprule
\textbf{Method} & \textbf{Top-k} & \textbf{Comp.} & \textbf{Rel.} & \textbf{Sim.}\\
\midrule
\multirow{3}{*}{RAG}
& 3 & 0.6547 & 0.6282 & 0.5662\\
& 6 & 0.6615 & 0.6259 & 0.5427\\
& 9 & 0.5809 & 0.5741 & 0.5582\\
\midrule
\multirow{3}{*}{CA-RAG}
& 3 & 0.6614 & 0.6306 & 0.5587\\
& 6 & 0.6742 & 0.6404 & 0.5615\\
& 9 & 0.6541 & 0.6350 & 0.5904 \\
\bottomrule
\end{tabular}
\end{table}

\subsubsection{Reproducibility}
\label{app:reproducibility}

All variants share the same retriever, base LLM, prompt scaffold, chunking strategy, decoding configuration, and evaluation pipeline. Retrieval uses identical corpora, embedding models, and top-$k$ settings across methods. Variants differ only in the source and type of structured constraints injected before retrieval and generation.

To minimise implementation variance, all experiments use deterministic prompting, fixed generation parameters, and identical clause-auditing protocols across domains and runs. Ontology matching and KG construction are largely automated from policy corpora and domain resources, reducing manual engineering requirements.

\subsection{Full Experimental Results}
\label{app:full_results}

This appendix presents full per-domain results for all evaluated models and generation variants, complementing the aggregated results reported in Section~\ref{sec:results}. We also define all abbreviated metric names that appear in the tables (e.g., Clause Cov.\ for DSA coverage score, Lat.\ for latency, TFLOPs/s for effective throughput, Sim. for similarity, Comp. for compliance score), covering the efficiency, compliance, and similarity dimensions. Unless otherwise stated, all results are reported without RLHF refinement.

\begin{table*}[ht]
\centering
\scriptsize
\caption{Detailed evaluation results on the \textbf{Health} domain (without RLHF).}
\label{tab:health_full}

\setlength{\tabcolsep}{5pt}

\begin{tabular}{l l c c c c c c}
\toprule
\textbf{Model} &
\textbf{Variant}
& \textbf{Tokens}
& \textbf{Lat (s)}
& \textbf{TFLOPs/s}
& \textbf{Clause Cov. (\%)}
& \textbf{Comp.}
& \textbf{Sim.}
\\
\midrule
\multirow{6}{*}{Mistral-7B} & RAG        & 1247 & 21.60 & 0.42 & 74.57 & 0.6731 & 0.6569 \\
           & RAG+O      & 1363 & 30.27 & 0.39 & 76.30 & 0.6604 & 0.6526 \\
           & RAG+KG     & 1433 & 31.57 & 0.43 & 78.93 & 0.6490 & 0.6574 \\
           & CA-RAG   & 1555 & 27.18 & 0.73 & 78.07 & 0.6727 & 0.6629  \\
           & \textsc{Self-RAG} & 3236 & 48.61 & 2.19 & 76.30 & 0.6663 & 0.6537 \\
           & \textsc{Flare-RAG} & 1603 & 53.73 & 0.47 & 79.83 & 0.6802 & 0.6727 \\
\midrule
\multirow{6}{*}{Llama-3.1-8B} & RAG  & 1267 & 16.83 & 0.63 & 79.80 & 0.6479 & 0.6750 \\
            & RAG+O & 1380 & 17.10 & 0.66 & 80.70 & 0.6499 & 0.6699 \\
            & RAG+KG & 1448 & 18.83 & 0.62 & 82.43 & 0.6143 & 0.6731 \\
            & CA-RAG & 1577 & 16.90 & 0.68 & 80.70 & 0.6453 & 0.6772 \\
            & \textsc{Self-RAG} & 3343 & 24.16 & 2.68 & 80.70 & 0.6649 & 0.6707 \\
           & \textsc{Flare-RAG} & 1926 & 24.67 & 1.45 & 77.17 & 0.6845 & 0.6670 \\
\midrule
\multirow{6}{*}{Llama-3.2-3B} & RAG & 1256 & 19.51 & 0.64 & 75.43 & 0.6652 & 0.6593 \\
             & RAG+O & 1356 & 16.34 & 0.71 & 78.90 & 0.6586 & 0.6676 \\
             & RAG+KG & 1387 & 15.78 & 0.85 & 77.17 & 0.6489 & 0.6140 \\
             & CA-RAG & 1556 & 17.83 & 0.91 & 78.07 & 0.6824 & 0.6512 \\
             & \textsc{Self-RAG} & 3196 & 28.44 & 4.32 & 73.67 & 0.6637 & 0.6521 \\
             & \textsc{Flare-RAG} & 1906 & 42.17 & 1.06 & 80.70 & 0.6554 & 0.6722 \\
\bottomrule
\end{tabular}
\end{table*}

\begin{table*}[ht]
\centering
\scriptsize
\caption{Detailed evaluation results on the \textbf{Finance} domain (without RLHF).}
\label{tab:finance_full}

\setlength{\tabcolsep}{5pt}

\begin{tabular}{l l c c c c c c}
\toprule
\textbf{Model} &
\textbf{Variant}
& \textbf{Tokens}
& \textbf{Lat (s)}
& \textbf{TFLOPs/s}
& \textbf{Clause Cov. (\%)}
& \textbf{Comp.}
& \textbf{Sim.}
\\
\midrule
\multirow{6}{*}{Mistral-7B} & RAG        & 1668 & 18.37 & 0.70 & 74.57 & 0.6267 & 0.5546 \\
           & RAG+O      & 1709 & 16.73 & 0.88 & 74.57 & 0.6878 & 0.5229 \\
           & RAG+KG     & 1957 & 19.07 & 0.97 & 78.07 & 0.6577 & 0.5214 \\
           & CA-RAG   & 2118 & 20.05 & 0.99 & 74.57 & 0.6780 & 0.5277  \\
           & \textsc{Self-RAG} & 4425 & 45.80 & 2.57 & 69.30 & 0.6377 & 0.5472 \\
           & \textsc{Flare-RAG} & 2374 & 58.17 & 0.73 & 71.97 & 0.6384 & 0.5482 \\
\midrule
\multirow{6}{*}{Llama-3.1-8B} & RAG  &  1677 & 18.07 & 0.78 & 75.43 & 0.6615 & 0.5427 \\
            & RAG+O & 1764 & 20.37 & 0.63 & 70.20 & 0.6442 & 0.5613 \\
            & RAG+KG & 1980 & 24.18 & 0.63 & 74.57 & 0.6571 & 0.5582 \\
            & CA-RAG & 2064 & 22.00 & 0.69 & 74.57 & 0.6742 & 0.5615 \\
            & \textsc{Self-RAG} & 4619 & 48.03 & 2.61 & 76.30 & 0.6163 & 0.5700 \\
           & \textsc{Flare-RAG} & 2745 & 42.40 & 1.30 & 74.53 & 0.6405 & 0.5537 \\
\midrule
\multirow{6}{*}{Llama-3.2-3B} & RAG & 1668 & 13.43 & 0.97 & 71.07 & 0.6442 & 0.5490 \\
            & RAG+O & 1745 & 13.40 & 0.92 & 71.07 & 0.6384 & 0.5468 \\
            & RAG+KG & 1932 & 14.53 & 0.99 & 74.53 & 0.6357 & 0.5527 \\
            & CA-RAG & 2064 & 12.77 & 1.36 & 78.07 & 0.6734 & 0.5653 \\
            & \textsc{Self-RAG} & 4473 & 26.63 & 4.41 & 71.03 & 0.6421 & 0.5774 \\
           & \textsc{Flare-RAG} & 2410 & 31.33 & 1.33 & 68.43 & 0.6320 & 0.5561 \\
\bottomrule
\end{tabular}
\end{table*}

\begin{table*}[ht]
\centering
\scriptsize
\caption{Detailed evaluation results on the \textbf{Education} domain (without RLHF).}
\label{tab:education_full}

\setlength{\tabcolsep}{5pt}

\begin{tabular}{l l c c c c c c}
\toprule
\textbf{Model} &
\textbf{Variant}
& \textbf{Tokens}
& \textbf{Lat (s)}
& \textbf{TFLOPs/s}
& \textbf{Clause Cov. (\%)}
& \textbf{Comp.}
& \textbf{Sim.}
\\
\midrule
\multirow{6}{*}{Mistral-7B} & RAG & 1253 & 20.04 & 0.39 & 77.20 & 0.6705 & 0.6316 \\
        & RAG+O & 1395 & 19.75 & 0.44 & 73.70 & 0.6911 & 0.6240 \\
        & RAG+KG & 1549 & 21.15 & 0.49 & 77.17 & 0.6551 & 0.5992 \\
        & CA-RAG & 1694 & 22.85 & 0.50 & 76.30 & 0.6798 & 0.6313  \\
        & \textsc{Self-RAG} & 3228 & 16.36 & 4.14 & 78.03 & 0.6869 & 0.6216  \\
        & \textsc{Flare-RAG} & 1912 & 38.65 & 0.99 & 70.17 & 0.6670 & 0.6115  \\
\midrule
\multirow{6}{*}{Llama-3.1-8B} & RAG & 1302 & 12.68 & 0.66 & 76.30 & 0.6748 & 0.6539 \\
             & RAG+O & 1455 & 10.82 & 0.87 & 80.70 & 0.6339 & 0.6625 \\
            & RAG+KG & 1539 & 11.00 & 0.92 & 81.57 & 0.6988 & 0.6593 \\ 
            & CA-RAG & 1732 & 13.34 & 0.91 & 81.57 & 0.6898 & 0.6671 \\
            & \textsc{Self-RAG} & 3427 & 66.38 & 1.30 & 85.93 & 0.6857 & 0.6574 \\
            & \textsc{Flare-RAG} & 2018 & 62.47 & 0.70 & 80.67 & 0.6607 & 0.6670 \\
\midrule
\multirow{6}{*}{Llama-3.2-3B} & RAG & 1282 & 08.01 & 1.04 & 78.93 & 0.6498 & 0.6554 \\
            & RAG+O & 1435 & 08.66 & 1.11 & 78.03 & 0.6813 & 0.6140 \\
            & RAG+KG & 1578 & 08.75 & 1.40 & 78.07 & 0.6607 & 0.6419 \\
            & CA-RAG & 1735 & 10.34 & 1.23 & 78.07 & 0.6579 & 0.6386 \\
            & \textsc{Self-RAG} & 3139 & 28.84 & 2.34 & 67.57 & 0.6335 & 0.5775 \\
            & \textsc{Flare-RAG} & 1915 & 46.04 & 0.76 & 71.07 & 0.6557 & 0.6230 \\

\bottomrule
\end{tabular}
\end{table*}

\subsection{Efficiency and Scalability}
\label{app:efficiency}

Table~\ref{tab:efficiency} reports latency and empirical throughput across retrieval variants. Throughput is reported as empirical GPU TFLOPs/s measured during inference under fixed hardware settings. CA-RAG introduces moderate overhead from constraint extraction and query construction, but remains substantially more efficient than adaptive retrieval baselines such as \textsc{Self-RAG} and \textsc{Flare-RAG}, which require iterative retrieval and refinement. \textsc{Self-RAG} and \textsc{Flare-RAG} generally incur higher latency than standard RAG and \textsc{CA-RAG}, although the magnitude varies across model--domain settings. In contrast, \textsc{CA-RAG} achieves improved omission robustness with comparatively bounded runtime overhead relative to standard \textsc{RAG}. These results suggest that retrieval-stage structural conditioning improves structural completeness without requiring expensive iterative retrieval loops, making CA-RAG practical for long-form constrained generation settings.

\begin{table}[t]
\centering
\scriptsize
\caption{Latency and empirical throughput (TF/s) across domains and retrieval variants. Latency (Lat.) is reported in seconds.}
\label{tab:efficiency}

\setlength{\tabcolsep}{3pt}
\renewcommand{\arraystretch}{0.85}

\begin{tabular}{l l cc cc cc}
\toprule
&
& \multicolumn{2}{c}{\textbf{Health}}
& \multicolumn{2}{c}{\textbf{Finance}}
& \multicolumn{2}{c}{\textbf{Edu.}} \\
\cmidrule(lr){3-4}
\cmidrule(lr){5-6}
\cmidrule(lr){7-8}

\textbf{LLM} &
\textbf{Method}
& Lat.
& TF/s
& Lat.
& TF/s
& Lat.
& TF/s \\
\midrule

\multirow{4}{*}{Mistral-7B}
& RAG & 21.6 & 0.42 & 18.4 & 0.70 & 20.0 & 0.39 \\
& CA-RAG & 27.2 & 0.73 & 20.1 & 0.99 & 22.8 & 0.50 \\
& \textsc{Self-RAG} & 48.6 & 2.19 & 45.8 & 2.57 & 16.4 & 4.14 \\
& \textsc{Flare-RAG} & 53.7 & 0.47 & 58.2 & 0.73 & 38.6 & 0.99 \\
\midrule

\multirow{4}{*}{Llama-3.1-8B}
& RAG & 16.8 & 0.63 & 18.1 & 0.78 & 12.7 & 0.66 \\
& CA-RAG & 16.9 & 0.68 & 22.0 & 0.69 & 13.3 & 0.91 \\
& \textsc{Self-RAG} & 24.2 & 2.68 & 48.0 & 2.61 & 66.3 & 1.30 \\
& \textsc{Flare-RAG} & 24.6 & 1.45 & 42.4 & 1.30 & 62.5 & 0.70 \\
\midrule

\multirow{4}{*}{Llama-3.2-3B}
& RAG & 19.5 & 0.64 & 13.4 & 0.97 & 8.0 & 1.04 \\
& CA-RAG & 17.8 & 0.91 & 12.8 & 1.36 & 10.3 & 1.23 \\
& \textsc{Self-RAG} & 28.4 & 4.32 & 26.6 & 4.41 & 28.8 & 2.34 \\
& \textsc{Flare-RAG} & 42.2 & 1.06 & 31.3 & 1.33 & 46.0 & 0.76 \\
\bottomrule
\end{tabular}
\end{table}

\subsection{Bootstrap Significance Testing}
\label{app:significance}

We performed paired bootstrap significance testing to evaluate whether improvements in document-level reliability were statistically significant across retrieval variants. Testing was conducted across domains, runs, and model variants using paired document-level reliability scores. Each bootstrap sample preserved document-level pairing between compared methods to isolate relative reliability differences under identical evaluation instances. For each comparison, we computed the mean paired difference and estimated $95\%$ confidence intervals using 5{,}000 bootstrap resamples.

Confidence intervals excluding zero were treated as statistically significant. Table~\ref{tab:bootstrap} shows that \textsc{CA-RAG} significantly outperformed standard \textsc{RAG}, \textsc{Self-RAG}, and \textsc{Flare-RAG}. In contrast, differences relative to the generation-only constraint baseline were not statistically significant, suggesting that generation-time structural prompting can partially improve aggregate reliability. However, the main paper shows that retrieval-stage structural conditioning more consistently improves omission robustness and clause-level completeness.
\begin{table}[t]
\centering
\small
\caption{Paired bootstrap significance testing using document-level reliability scores across domains, runs, and model variants. Confidence intervals excluding zero indicate statistically significant differences.}
\label{tab:bootstrap}
\setlength{\tabcolsep}{2pt}
\renewcommand{\arraystretch}{1.05}
\resizebox{\columnwidth}{!}{
\begin{tabular}{l c c c}
\toprule
\textbf{Baseline} 
& \textbf{Mean Diff.} 
& \textbf{95\% CI} 
& \textbf{Significant} \\
\midrule

RAG 
& +0.0122 
& [0.0034, 0.0207] 
& Yes \\

\textsc{Self-RAG}
& +0.0171 
& [0.0024, 0.0352] 
& Yes \\

\textsc{Flare-RAG}
& +0.0116 
& [0.0002, 0.0236] 
& Yes \\

\textsc{Gen-Constraint}
& +0.0099 
& [-0.0006, 0.0203] 
& No \\

\bottomrule
\end{tabular}
}
\end{table}
\subsection{Ablation: Impact of RLHF Refinement}
\label{sec:app_rlhf_ablation}

We analyse RLHF as a post-generation refinement stage while keeping retrieval and evidence support fixed. Table~\ref{tab:rlhf} reports reliability scores after RLHF refinement on \textbf{Llama-3.1-8B}. The effect of RLHF varies across domains and retrieval settings. In some cases, refinement improves stylistic coherence and local fluency, while in others it slightly reduces reliability, particularly in finance, where compliance-sensitive phrasing is more fragile.

Importantly, RLHF does not alter the relative behaviour of retrieval variants, nor does it consistently recover omitted obligations when supporting evidence is incomplete. Even when readability improves, omission-sensitive clauses such as transfer safeguards, legal-basis requirements, or compliance dependencies remain constrained by the quality and structural completeness of retrieved evidence. These results suggest that RLHF primarily acts as a lightweight stylistic refinement mechanism rather than a substitute for retrieval-stage structural alignment.

\begin{table}[t]
\centering
\small
\caption{RLHF ablation on \textbf{Llama-3.1-8B}. Values denote reliability scores after RLHF refinement, with relative change $\Delta$ in \%.}
\label{tab:rlhf}
\setlength{\tabcolsep}{2.5pt}

\begin{tabular}{l c c c}
\toprule
\textbf{Method} 
& \textbf{Health} 
& \textbf{Finance} 
& \textbf{Education} \\
\midrule

\textsc{RAG}      
& $0.6554 (-0.1)$ 
& $0.6152 (-1.7)$ 
& $0.6616 (-1.0)$ \\

\textsc{RAG+O}    
& $\mathbf{0.6805 (+3.8)}$ 
& $0.6103 (-1.5)$ 
& $0.6665 (+3.7)$ \\

\textsc{RAG+KG}   
& $0.6636 (+5.0)$ 
& $\mathbf{0.6300 (+0.4)}$ 
& $0.6555 (-3.8)$ \\

\textsc{CA-RAG}   
& $0.6604 (+0.8)$ 
& $0.6033 (-5.8)$ 
& $\mathbf{0.6744 (-1.7)}$ \\
\bottomrule
\end{tabular}
\end{table}

\subsection{Additional Structural and Compliance Trade-off Plots}
\label{app:radar}

\begin{figure*}[t]
  \centering
  \includegraphics[width=0.9\linewidth]{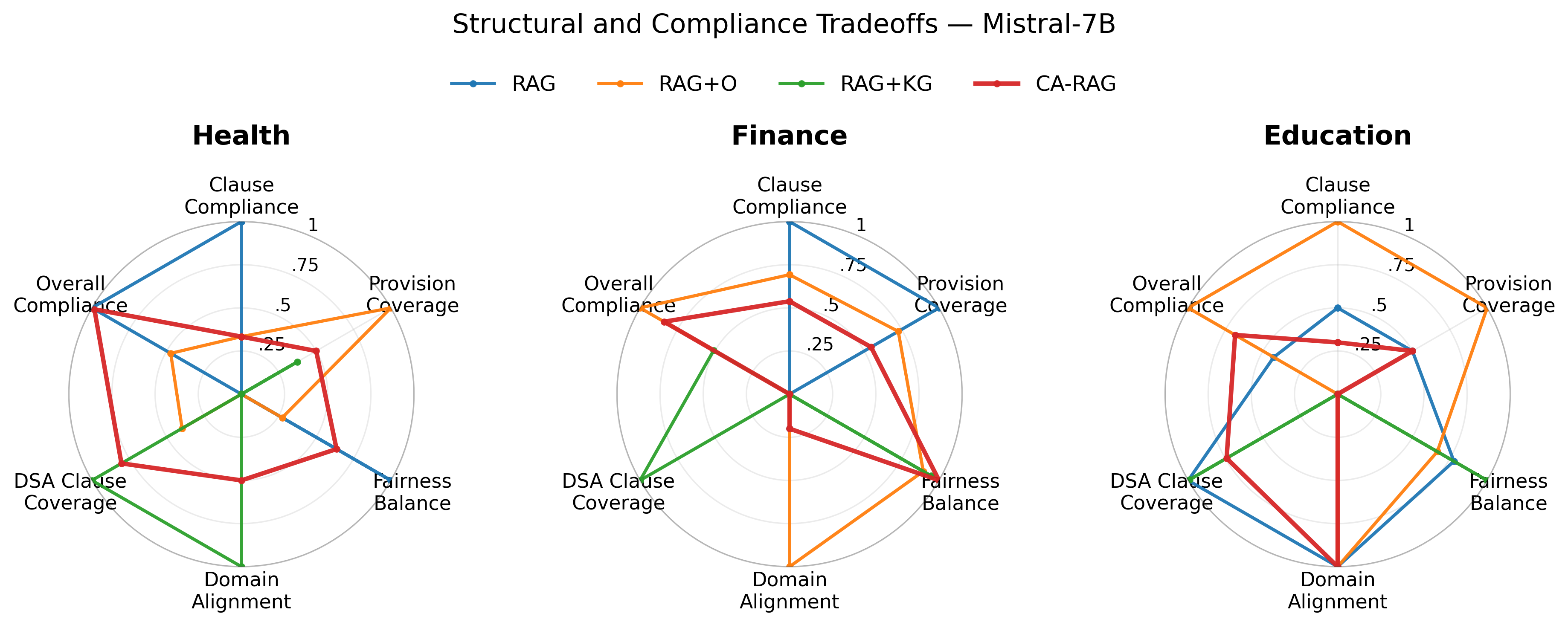}
  \caption{Structural and compliance tradeoffs across variants for \textbf{Mistral-7B}. Scores are normalised within each domain across six auditing metrics.}
  \label{fig:radar_mistral}
\end{figure*}

\begin{figure*}[t]
  \centering
  \includegraphics[width=0.9\linewidth]{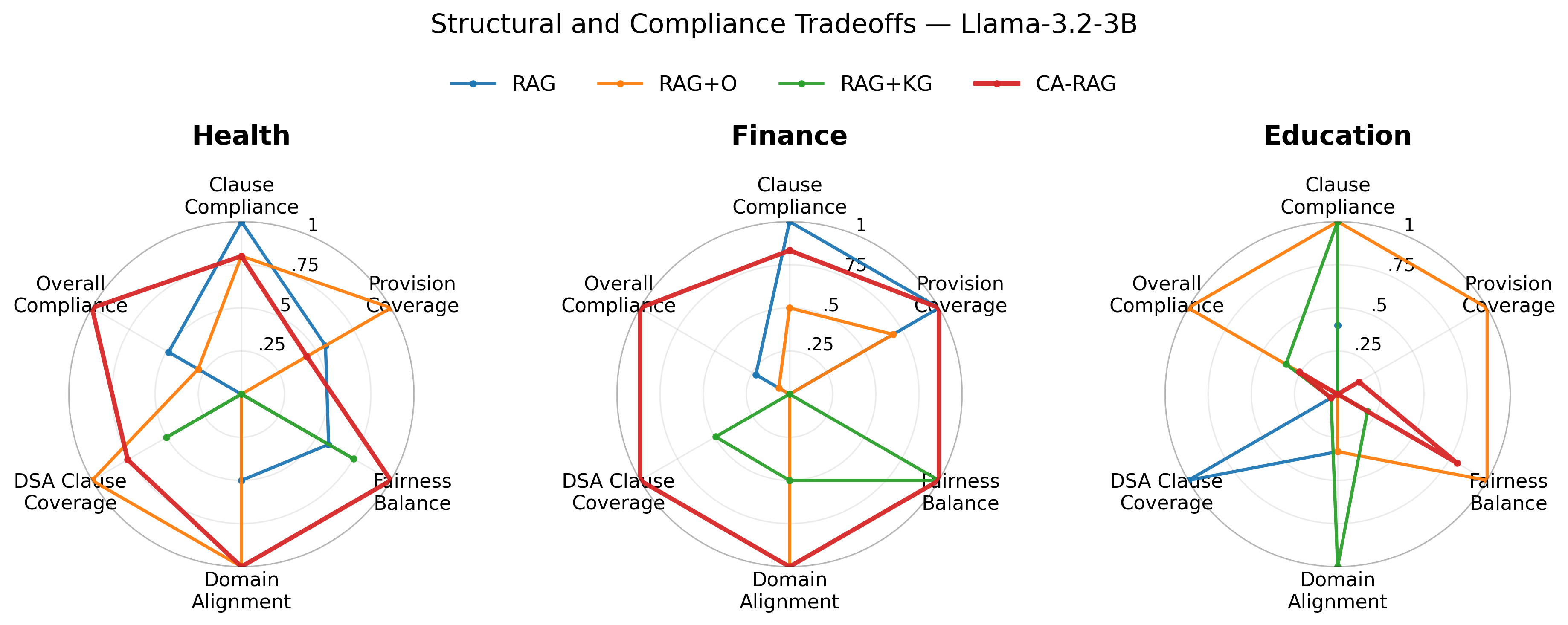}
  \caption{Structural and compliance tradeoffs across retrieval variants for \textbf{Llama-3.2-3B}. Scores are normalised within each domain across six auditing metrics.}
  \label{fig:radar_l32}
\end{figure*}

Figures~\ref{fig:radar_mistral} and~\ref{fig:radar_l32} provide additional radar plots for \textbf{Mistral-7B} and \textbf{Llama-3.2-3B}. These plots complement the main paper analysis by showing how retrieval variants trade off clause compliance, provision coverage, fairness and balance, domain alignment, DSA clause coverage, and overall compliance across domains.

\subsection{Qualitative Error Analysis}
\label{sec:qualitative}

We analyse omission failures across retrieval variants to examine retrieval-stage structural alignment. Table~\ref{tab:error_analysis} summarises representative omission patterns across finance, healthcare, and education DSAs. The High/Medium/Low labels provide descriptive summaries of omission patterns observed during qualitative inspection and should not be interpreted as quantitative frequency estimates. Transfer clauses include cross-border transfer obligations and data-sharing conditions; policy clauses include legal basis and compliance requirements; procedural clauses include incident response, audit, and review obligations. Common failures involve structurally dependent clauses such as \textit{Cross-Border Transfers}, \textit{Legal Basis}, \textit{DPIA/TIA\footnote{Transfer Impact Assessment.}}, \textit{Incident Response}, and \textit{Audit Requirements}. These obligations often depend on evidence distributed across multiple policy sections, making them difficult to recover through relevance-only retrieval. Standard \textsc{RAG} frequently produces semantically plausible but structurally incomplete outputs. For example, in \textit{Third-Party Transfer} section, standard RAG often retrieves general data-sharing policies while omitting safeguards such as SCC\footnote{Standard Contractual Clause.} or DPIA obligations. In contrast, \textsc{CA-RAG} retrieves evidence linked to transfer mechanisms and compliance dependencies, improving downstream clause recovery. 

Adaptive retrieval methods such as \textsc{Self-RAG} and \textsc{Flare-RAG} improve local coherence but still omit globally dependent obligations. Similarly, generation-only constraint injection improves structural phrasing but cannot reliably recover clauses unsupported by retrieved evidence. Overall, \textsc{CA-RAG} reduces omission frequency through retrieval-stage structural conditioning. Such omissions are particularly critical in compliance-sensitive settings, where missing obligations may matter more than small reductions in semantic similarity. Additional retrieval examples and omission traces are provided in Appendix~\ref{app:worked_examples}.

\begin{table}[t]
\centering
\footnotesize
\caption{Representative omission patterns across retrieval variants.}
\label{tab:error_analysis}
\setlength{\tabcolsep}{3pt}
\renewcommand{\arraystretch}{1.05}
\begin{tabular}{l c c c}
\toprule
\textbf{Method} 
& \textbf{Transfer} 
& \textbf{Policy} 
& \textbf{Procedural} \\
\midrule
RAG            & High   & High   & Medium \\
\textsc{Self-RAG}       & Medium & Medium & High \\
\textsc{Flare-RAG}      & Medium & Low    & Medium \\
\textsc{Gen-Constraint} & High   & High   & Medium \\
CA-RAG         & Low    & Low    & Low \\
\bottomrule
\end{tabular}
\end{table}
\end{document}